# Recovering Biomechanical Signals from Missing Keypoints Using Temporal Interpolation in Monocular Gait Analysis

Shubham Rajeshkumar Jariwala

Information Systems Technology and Design, Singapore University of Technology and Design, 8 Somapah Road, Singapore 487372, Singapore

ORCID: 0009-0005-4735-4759

*Corresponding author: Shubham Jariwala (email: shubhamrajeshkumar_jariwala@mymail.sutd.edu.sg)*

## Abstract

Monocular pose estimation enables low-cost gait analysis but is sensitive to missing keypoints caused by occlusion, detection errors, or efficiency-driven model reduction. While prior work on recovering missing joints focuses on complex learned models, the effectiveness of simple temporal methods remains underexplored. We evaluate knee-angle estimation under a missing-ankle-keypoint condition and test a first-order temporal interpolation scheme as a recovery mechanism. Across 527 frames of monocular walking video (428 with valid baseline detections), removing the ankle keypoint increased mean angular error to 23.4° ± 46.7° and collapsed signal variance to near zero. Temporal interpolation reduced error to 1.1° ± 6.7° (Wilcoxon signed-rank, $p < 10^{-50}$) and restored variance and smoothness to within a few percent of baseline. These results indicate that gait signals possess sufficient temporal redundancy for a simple, computationally trivial interpolation scheme to recover a critical missing joint, without resorting to learned reconstruction models. The findings support low-complexity, real-time-compatible designs for gait analysis in resource-constrained or occlusion-prone monocular settings.

## 1. Introduction

Monocular human pose estimation has become an attractive route to low-cost, camera-only gait analysis, avoiding the expense and lab constraints of marker-based motion capture [1,2]. Its practical value, however, depends on the reliability of individual keypoint estimates, which degrade under occlusion, motion blur, low-confidence detections, or the use of lighter models chosen for real-time throughput.

Existing approaches to recovering missing pose information generally rely on learned priors. These include Bayesian and geometric-constraint models that infer occluded joints directly [7,8], temporal-context networks that exploit information across frames [3,9], and full 3D reconstruction pipelines such as OpenCap [4] or deep learning-based monocular capture systems [5], which fuse multi-view or biomechanically constrained information.

It remains unclear whether a much simpler approach treating the missing keypoint purely as a temporal gap and filling it from adjacent frames is sufficient to recover a biomechanically meaningful signal. This is the question addressed here, using knee-angle estimation as the target signal and ankle-keypoint loss as the perturbation, since the ankle is both commonly occluded in monocular views and directly load-bearing for the knee-angle calculation.

## 2. Methods

### 2.1 Pipeline

Monocular walking video (a single participant, male, height 1.70 m, mass 64 kg) was processed frame-by-frame with a YOLO pose estimator (YOLOv8-pose family) [6] to extract the 17 COCO-format keypoints [10]. The knee angle θ was computed from the hip (H), knee (K), and ankle (A) keypoints as the angle at K between vectors K→H and K→A:

$$\theta = \arccos\left[ (K\to H \cdot K\to A) / (\|K\to H\| \, \|K\to A\|) \right] \quad (1)$$

### 2.2 Conditions

Three conditions were compared over the same 527-frame sequence: (i) full baseline, knee angle computed from all detected keypoints; (ii) missing ankle, the ankle keypoint discarded for every frame, simulating a hard occlusion or detector dropout; and (iii) temporal interpolation, in which the missing ankle position is recovered using a first-order exponential smoothing update, $\text{ankle}'_t = 0.7 \cdot \text{ankle}'_{t-1} + 0.3 \cdot \text{ankle}_t$ (using the raw detected position directly when no prior recovered value exists), and the knee angle is recomputed from this smoothed position.

Even in the full-baseline condition, YOLO failed to return valid keypoints for 99 of 527 frames (baseline detection rate: 81.2%), reflecting realistic detector dropout on a single monocular recording rather than an idealized signal. All angular-error and statistical comparisons below use the 428 frames with valid baseline detections, so that recovery performance is measured against a real, rather than synthetic, reference signal.

### 2.3 Metrics and statistics

For each condition we computed: (i) mean angular error relative to the full-baseline knee angle, (ii) signal variance, and (iii) signal smoothness, defined as the mean absolute frame-to-frame change, $\text{mean}(|\theta_t - \theta_{t-1}|)$. Missing-ankle versus interpolated errors were compared using the Wilcoxon signed-rank test, chosen over a paired t-test because per-frame error distributions are right-skewed, with standard deviations exceeding their means.

The video recording used in this study was of the author, who provided consent for its use in this research. No third-party participant data were collected.

## 3. Results

Table 1 summarizes mean angular error, standard deviation, variance, and smoothness for the three conditions, computed over the 428 frames with valid baseline detections.

| Condition | Mean error (°) | SD (°) | Variance | Smoothness |
|---|---|---|---|---|
| Full baseline | - | - | 2177.8 | 1.97 |
| Missing ankle | 23.4 | 46.7 | ~0.0 | 0.01 |

| Condition | Mean error (°) | SD (°) | Variance | Smoothness |
|---|---|---|---|---|
| Interpolated | 1.1 | 6.7 | 2192.0 | 2.12 |

**Table 1. Angular error, variance, and smoothness across the full baseline, missing-ankle, and interpolated conditions (n = 428 valid frames).**

Removing the ankle keypoint degrades the signal on all three measures simultaneously: error rises sharply, and variance and smoothness both collapse toward zero, indicating the knee-angle trace becomes effectively frozen once the ankle is lost consistent with the angle calculation losing one of its two defining vectors, rather than merely becoming noisier.

Temporal interpolation reverses this collapse rather than only reducing the point-wise error: variance and smoothness after interpolation both return to within a few percent of baseline, alongside the large reduction in mean error (Wilcoxon signed-rank test, missing-ankle vs. interpolated, $p \approx 7.6 \times 10^{-59}$). This indicates that the recovered signal is not just closer to baseline on average, but restores the underlying temporal dynamics of the gait cycle.

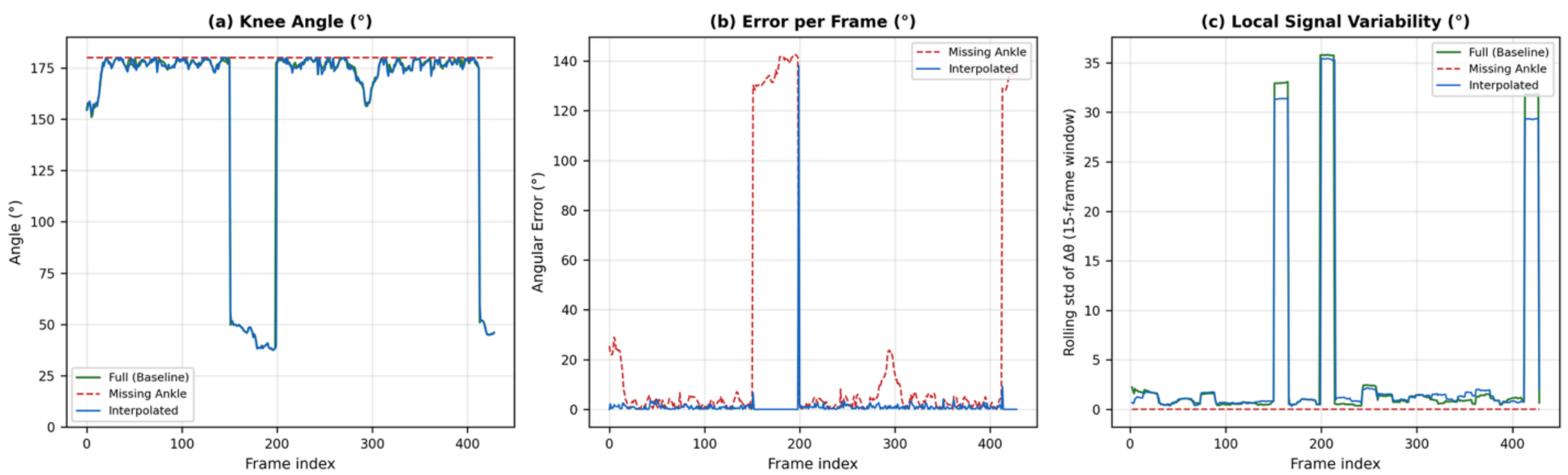

Figure 1: pipeline diagram and knee-angle trajectory / error / smoothness comparison across conditions.

## 4. Discussion

These results show that a first-order temporal interpolation arguably the simplest possible recovery strategy is sufficient to recover a biomechanically meaningful knee-angle signal after losing a keypoint that is directly load-bearing for the computation. This suggests that gait signals carry strong short-timescale redundancy: the ankle's position over a single missing frame is well predicted by its immediately preceding position, at typical video frame rates and walking cadences.

This has a practical implication for monocular gait-analysis system design: robustness to occasional keypoint dropout (occlusion, low-confidence detections, or the 81.2% baseline detection rate observed even without any injected perturbation in this dataset) does not necessarily require a learned reconstruction model. A negligible-cost interpolation step can substantially mitigate this class of failure.

## 5. Limitations

This study uses a single participant and a single walking sequence, so the generalizability of the recovery magnitude (23.4° → 1.1°) to other gait patterns, camera angles, or occlusion durations is untested. The interpolation scheme was evaluated only for single-frame ankle loss; performance under longer occlusion runs (multiple consecutive missing frames) is expected to degrade and was not characterized here. All reference ("ground truth") angles are themselves derived from the same monocular pipeline's own detections rather than an independent motion-capture reference, so absolute error magnitudes should be interpreted as internal consistency rather than validated accuracy against a gold standard.

## 6. Conclusion

Simple temporal interpolation recovers a knee-angle signal from a missing, load-bearing keypoint with high fidelity (23.4° → 1.1° mean error) in a monocular gait pipeline, without requiring a learned reconstruction model. This supports low-complexity designs for robust, real-time-compatible gait analysis in occlusion-prone or resource-constrained monocular settings.

## Acknowledgements

The author thanks the Agency for Science, Technology and Research (A*STAR), Singapore, for the opportunity to undertake this work during an attachment.

## CRediT author statement

Shubham Jariwala: Conceptualization, Methodology, Software, Formal analysis, Investigation, Data curation, Writing – original draft, Writing – review & editing, Visualization.

## Data availability statement

The analysis code and experiment CSV files supporting this study will be made publicly available in a GitHub repository upon acceptance of this manuscript. Prior to acceptance, the repository is maintained privately by the corresponding author to protect priority of the findings. The monocular video recording is not publicly shared to protect participant identifiability; it is available from the corresponding author on reasonable request.

## Declaration of competing interests

The author declares no competing financial interests or personal relationships that could have appeared to influence the work reported in this paper.